# MATHEMATICAL FOUNDATIONS FOR DESIGNING AND DEVELOPMENT OF INTELLIGENT SYSTEMS OF INFORMATION ANALYSIS

*D.O. Terletskyi[1], O.I. Provotar[2]*

Taras Shevchenko National University of Kyiv, Cybernetics Faculty

03680, Kyiv, Academician Glushkov Avenu 4d,
dmytro.terletskyi@gmail.com[1], aprowata@unicyb.kiev.ua[2]

This article is an attempt to combine different ways of working with sets of objects and their classes for designing and development of artificial intelligent systems (AIS) of analysis information, using object-oriented programming (OOP). This paper contains analysis of basic concepts of OOP and their relation with set theory and artificial intelligence (AI). Process of sets and multisets creation from different sides, in particular mathematical set theory, OOP and AI is considered. Definition of object and its properties, homogeneous and inhomogeneous classes of objects, set of objects, multiset of objects and constructive methods of their creation and classification are proposed. In addition, necessity of some extension of existing OOP tools for the purpose of practical implementation AIS of analysis information, using proposed approach, is shown.

Дана стаття є спробою поєднати різні підходи до роботи з множинами об’єктів та їхніми класами задля проектування і розробки систем штучного інтелекту (СШІ) для аналізу інформації, використовуючи об’єктно-орієнтоване програмування (ООП). Робота містить аналіз базових концепцій ООП і їхній зв'язок з теорією множин та штучним інтелектом (ШІ). Розглядається процес створення множин та мультимножин з різних сторін, зокрема з огляду на математичну теорію множин, ООП та ШІ. Запропоновані визначення об’єкту та його властивостей, однорідного та неоднорідного класу об’єктів, множини об’єктів, мультимножини об’єктів і конструктивні методи їх створення та класифікації. Окрім того, показана необхідність в деякому розширенні засобів ООП з метою практичної реалізації СШІ для аналізу інформації, використовуючи запропонований підхід.

## Introduction

Modern programming includes many different paradigms, approaches, techniques and programming languages. Object-Oriented Programming (OOP) is one of the famous and useful programming paradigm nowadays. Indeed, according to [1-3] the most popular programming languages are such languages, which support OOP.

As we know, *object* and *class* are the main concepts of OOP. According to [4], objects are the building blocks of an object-oriented program. We associate these blocks with the objects of real world, during developing programs. Concerning classes, they are blueprints, which we use as the basis for objects building. Every object is defined by two terms: attributes and behaviors. Attributes are properties of object, which describe it, and behaviors are procedures, functions (methods) which we can apply to this object and change its state, form and so on. According to [5], real world is created by objects, and OOP is the approach for description and simulation of this world or some his particular parts.

It is obvious, if person can develop the programs, which operate with models of objects of real world, their classes and so on, then this person precisely knows how to operate with them using his mind (intellect). Of course, sometimes people do this consciously, sometimes not, but it is important that they do this, and simple proof of this fact is that people have invented OOP. However, real world is very complicated, and consists of huge number of objects of different classes, that is why formalization of objects of real world into OOP objects and classes sometimes is nontrivial. Despite this, programming, in particular OOP, is developing day by day. People are finding more and more new practical applications for programming, and such area as Artificial Intelligence (AI) is not exception.

Modern AI includes many directions as knowledge representation, automated planning and scheduling, machine learning, robotics, computer vision, artificial creativity, natural language processing, and so on. Nevertheless, all these directions have something common among themselves. One of the main targets for all these directions is development of intelligent information systems (IIS) or artificial intelligent systems (AIS) for solving particular practical problems in these areas. Also very important aspect of AIS is intellectual analysis of information, in particular in real world or in some its parts. Ideologically this is very close to process of development of object-oriented programs. However, software development which is just tool for solving particular tasks, management by some process, etc. is easier then development of software which has some level of individuality and intellectuality and can do something bigger than just some computations. That is why questions about usability of development AIS using OOP approach appear.

Nowadays very often development of AIS is reduced to heuristic programming [6]. The advantage is that such approach gives an opportunity to solve corresponding practical problems, but the disadvantage is that partial solution does not guarantees general solution of corresponding problems. However, every AIS is based on some model of knowledge representation (KR). Nowadays, there is a variety of them. The most famous and common are Semantic Nets, Frames [7-13] and Logical Models [9-14] also Scripts [8-10], Conceptual Dependency [8, 10], Search Spaces, Search Trees [7], Petri Nets [8], K-Lines, Memory Organization Packets [9], Semantic WEB, Computational

Knowledge Discovery [11] are known. Furthermore, there are a few hybrid models of KR, in particular Brooks' Subsumption Architecture and Copycat [10].

Each of these models has own specifics and is useful in the particular domain. However, during developing certain AIS we need to implement a model of KR. Therefore, in this case, we have at least two levels of our AIS, first of them is a level of KR model and second one is a level of its practical implementation. In some cases, implementation of particular KR model can cause additional problems and difficulties. That is why, sometimes it leads to development of new programming paradigms and languages, which are invented for effective implementations of particular KR model. One of such examples is programming language PROLOG, which is logical programming language and it gives us opportunities for implementations of different Logical models of KR.

One of the main postulates of OOP is that objects create world. Development of OO-programs is formal description of objects from some world and relations between them, etc. That is why we can conclude that OOP can be one of the KR models, which represents (describes) knowledge about objects. Such model is very close to human perception of real world, because a person perceives real world by objects of this world. This fact gives us an opportunity to combine level of KR model and level of its practical implementation, during AIS development. In addition, a person also uses concept of class and set of objects in the process of thinking. Really, we use classes and sets in our mental activity during perception, analysis, comparison, retrieval, classification every day. We create classes and sets consciously or subconsciously, operate with them and apply a variety of operations to them. Moreover, set is the central concept in set theory and one of the most important for mathematics in general. However, questions about the origin of specific sets are emerging while analyzing the definition of this term, which is given in [15]. We can conclude that the “new” set can be obtained by set-theoretic operations over “existing” sets, analyzing the different systems of set theory [16, 17], and it is really so. However, the questions about origin of these so-called “existing” sets, their number, their types and so on do not disappear. Our target is development of AIS for intellectual analysis of information, based on operating with sets of objects and their classes, using OOP. However, primarily we need to understand specifics of classes' and sets creation from different sides, i.e. sets in the set theory, sets and classes in OOP and in the process of human thinking.

## Objects and Classes

**Objects.** We know that each set consists of elements, which form it. Everything, phenomena of our imagination or of our world can be the elements of the set. It is convenient for us to call them objects. Let consider such object as “natural number”. It is clear that every natural number must be integer and positive. These are characteristic properties of natural numbers. It is obvious, that 11 is really a natural number, but −16 and 9.52, for example, are not natural numbers. We can conclude that each object has certain properties, which define it as some essence while analyzing this fact. Usually in OOP [4, 5], we can consider properties of objects separately from objects. Such form of consideration is very close to concept of object-oriented class (OO class). Actually, objects and their properties cannot exist separately, because if we assume the opposite, we will have contradiction. On the one hand, object cannot exists separately from its properties, because without properties we cannot imagine and cannot describe it. On the other hand, object's properties cannot exist separately from object, because without object we cannot see and cannot perceive them. That is why, we cannot consider them separately, and there are few variants of the definitions order. It means that we cannot introduce definition of object without definition of its properties and vice versa. Therefore, we decided to introduce concept of object's properties firstly.

Globally we can divide properties of objects into two types – quantitative and qualitative. We will define these two types of object's properties formally, but their semantics has intuitive nature.

**Definition 1.** *Quantitative property of object* $A$ *is a tuple* $p_i(A) = (v(p_i(A)), u(p_i(A)))$, *where* $i = \overline{1,n}$, $v(p_i(A))$ *is an quantitative value of* $p_i(A)$, *and* $u(p_i(A))$ *are units of measure of quantitative value of* $p_i(A)$.

Let us consider some examples of quantitative properties of objects. Suppose we have an apple, and one of its properties is weight. We can present this property as follows $p_{weigth}(Apple) = (v(p_{weigth}(Apple)), u(p_{weigth}(Apple)))$, and if our apple has the weight of 0.2 kg, then property $p_{weight}(Apple)$ will be the following $p_{weight}(Apple) = (0.2, kg)$. Suppose we have a phone number, and one of its properties is sequence of odd numbers, which it comprises. We can present this property as follows $p_{oddnum}(PhoneNumber) = (v(p_{oddnum}(PhoneNumber)), u(p_{oddnum}(PhoneNumber)))$, and if your phone number includes following odd numbers 3,7,5, then property $p_{oddnum}(PhoneNumber)$ will be the following $p_{oddnum}(PhoneNumber) = ((3,7,5), numbers)$.

**Definition 2.** *Two quantitative properties* $p_i(A)$ *and* $p_j(B)$, *where* $i = \overline{1,n}$, $j = \overline{1,m}$, *are equivalent, i.e.* $Eq(p_i(A), p_j(B)) = 1$, *if and only if* $u(p_i(A)) = u(p_j(B))$.

**Definition 3.** *Qualitative property of object* $A$ *is a verification function* $p_i(A) = vf_i(A)$, $i = \overline{1,n}$ *which defines as a mapping* $vf_i(A): p_i(A) \rightarrow [0,1]$.

Let us consider some examples of qualitative properties of objects. Suppose we have an integer number $n$, and one of its properties is positivity. We can present this property as follows $p_{positivity}(n) = vf_{positivity}(n)$, where $vf(n)$ is verification function of property $p_{positivity}(n)$. In this case, function $vf_{positivity}(n): p_{positivity}(n) \to \{0,1\}$, and it is a particular case of verification function – predicate or Boolean-valued function.

Let us consider such object as car, and one of its properties is “high speed”. Suppose that maximum speed of this car is 200 km /hour. In this case we cannot precisely conclude about $vf_{highspeed}(car)$, because “high speed” is fuzzy concept [18]. There are many ways to define value of $vf_{highspeed}(car)$, and everything depends on definition of high speed. For example if high speed is 150 km/hour, then we can conclude that $vf_{highspeed}(150km/hour) = 1$, and $vf_{highspeed}(0km/hour) = 0$. Based on this and using proportion we can conclude that $vf_{highspeed}(75km/hour) = 0.5$.

We can conclude that, such approach gives an opportunity to combine description of property and its verification in the one function, i.e. verification function is a verification function and a description of property at the same time. Therefore, different algorithms can be verifiers and descriptors of property simultaneously.

**Definition 4.** *Two qualitative properties* $p_i(A)$ *and* $p_j(B)$, *where* $i = \overline{1,n}$, $j = \overline{1,m}$, *are equivalent, i.e.* $Eq(vf_i(A), vf_j(B)) = 1$, *if and only if* $(vf_i(A) = vf_j(A)) \wedge (vf_i(B) = vf_j(B))$.

**Definition 5.** *Specification of object* $A$ *is a vector* $P(A) = (p_1(A),...,p_n(A))$, *where* $p_i(A)$, $i = \overline{1,n}$ *is quantitative or qualitative property of object* $A$.

**Definition 6.** *Dimension of object* $A$ *is number of properties of object* $A$, *i.e.* $D(A) = |P(A)|$.

Now, we can formulate the definition of “object”.

**Definition 7.** *Object is a pair* $A / P(A)$, *where* $A$ *is object’s identifier and* $P(A)$ *– specification of object.*

Essentially, object is a carrier of some properties, which define it as some essence.

**Definition 8.** *Two objects* $A_1$ *and* $A_2$ *are similar, if and only if* $P(A_1) = P(A_2)$.

**Classes of Objects.** In general, we can divide objects on concrete and abstract, and does not matter when or how someone created each particular object. It is material implementation of its abstract image – *a prototype*. This prototype is essentially an abstract specification for creation the future real objects. Besides properties of objects, we should allocate *operations (methods)* which we can apply to objects, considering the features of their specifications. Really, we can apply some operations (methods) to objects for their changing and for operating with them. That is why, it will be useful to define concept of object's operation (method).

**Definition 9.** *Operation (method) of object* $A$ *is a function* $f(A)$, *which we can apply to object* $A$ *considering the features of its specification.*

For example, for such objects as natural numbers $n$, $m$ we can define operations “+” and “*”.

In OOP [4, 5, 19, 20], programmers operate with specifications and methods of objects without objects, and they call it *a type* or *a class* of object. It consists of fields and methods. Fields of class, essentially, are specification of class. Methods are functions, which we can apply to objects of this class for their changing and for operating with them. Concept of OO class is similar to universal algebra [21, 22], where carrier of algebra consists of objects and signature consists of methods of objects. That is why, henceforth sometimes we will use title signature of class for methods of class. Let us define concept of object's signature.

**Definition 10.** *Signature of object* $A$ *is a vector* $F(A) = (f_1(A),...,f_m(A))$, *where* $f_i(A)$, $i = \overline{1,m}$ *is an operation (method) of object* $A$.

Generally, signature of particular object can consist of different quantity of operations, but in practice, especially in programming, usually we are considering finite signatures of objects.

According to definition of object, every object has some specification, which defines it as some essence. There are some objects, which have similar specifications. It means that we can apply the same methods to them. Let us define similar objects.

**Definition 11.** *Objects A and B are similar objects, if and only if, when they have the same dimension and equivalent specifications.*

If certain two objects are similar, we can conclude that these objects have the same type or class. Now we can introduce concept of object's class.

**Definition 12.** *Object’s class* $T$ *is a tuple* $T = (P(T), F(T))$, *where* $P(T)$ *is abstract specification of some quantity of objects, and* $F(T)$ *is their signature.*

When we talk about class of objects, we mean properties of these objects and methods, which we can apply to them. Class of objects is a generalized form of consideration of objects and operations on them, without these objects.

As an example, let us describe type *Int* in programming language C++, using concept of similar objects and object’s class. Let us set the next specification for class *Int*

$$P(Int) = (p_1(Int), p_2(Int)),$$

where property $p_1(Int)$ means “integer number”, property $p_2(Int)$ means “number not bigger then 2147336147 and not smaller then −2147336148”. It is obvious, that all numbers which have properties $p_1(Int)$ and $p_2(Int)$ are objects of class $Int$. Let define the methods of class $Int$ in the following way:

$$F(Int) = (f_1(Int), f_2(Int)),$$

where $f_1(Int) = "+"$ and $f_2(Int) = "*"$.

Let us define concept of homogeneous class of objects.

**Definition 13.** *Homogeneous class of objects $T$ is a class of objects, which contains only similar objects.*

Considering concept of OO class, we can conclude that class is a prototype for particular objects, and all objects of class are described by class description. It means that every particular object of the class has the same specification and signature. That in turn, imposes some constraints on description of objects from real world. There are many different objects of real world, which belong to different classes, and if we need to work with them, we can describe them, using new class for each new type of objects. Especially, if we work with not very big quantity of different types of objects, we can do it without any fears. However, if we need to work with huge quantity of different types, for example, with a few thousands of different types, just a process of description of such types is very complex and time-consuming not to mention size of code and performance of such programs.

According to definitions of object and object’s class, we can conclude two points. Firstly, every object is a member of at least one class of objects, and secondly, objects and their classes cannot exist separately. Furthermore, some objects are members of few classes simultaneously. For example, such objects as natural numbers $n_1,\ldots,n_n$ are members of such classes as natural numbers, integer numbers, rational numbers and real numbers, i.e. $n_1,\ldots,n_n \in N \in Z \in Q \in R$. As we can see, class $R$ has the biggest cardinality in this case; furthermore, it consists of groups of objects of different types. It contradicts concept of OO class, because different objects from one OO class cannot have different specifications and signatures. That is why we cannot describe such class of objects using OO class.

Let consider the class of real numbers $R$, and describe it by the following specification $P(R) = (p_1(R),\ldots,p_5(R),\ldots)$, where $p_1(R)$ is “integer number”, $p_2(R)$ is “natural number”, $p_3(R)$ is “fractional number”, $p_4(R)$ is “negative integer”, $p_5(R)$ is “even number”. Let consider following numbers 3, 2.75, −16, 4, −7.48. It is obvious that they are objects of different types, but all of them are objects of class $R$. However, if these objects belong to class $R$, they must conform to the specification $P(R)$. Clearly that they do so, but in different ways (see Table 1).

**Table 1.** Conformity of objects 3, 2.75, −16, 4, −7.48 to the class $R$.

| $p_j(A_i)/A_i$ | 3 | 2.75 | −16 | 4 | −7.48 |
|---|---|---|---|---|---|
| Integer number | 1 | 0 | 1 | 1 | 0 |
| Natural number | 1 | 0 | 0 | 1 | 0 |
| Fractional number | 0 | 1 | 0 | 0 | 1 |
| Negative integer | 0 | 0 | 1 | 0 | 1 |
| Even number | 0 | 0 | 1 | 1 | 1 |

It is obvious, that mathematically all these numbers 3, 2.75, −16, 4, −7.48 are objects of the class $R$, but, as we can see, we really cannot describe them using one OO class. Of course, real numbers implemented for many program languages as one of the primitive types of data, which we can use without any descriptions, because it is built-in. However, it is one of the examples of inhomogeneous class of objects.

As we can see, there are two different types of object classes, that is why, let us define concept of inhomogeneous class of objects.

**Definition 14.** *Inhomogeneous class of objects $T$ is a tuple $T = (Core(T), pr_1(A_1),\ldots,pr_n(A_n))$, where $Core(T) = (P(T), F(T))$ is the core of class $T$, which includes properties and methods similar to specifications $P(A_1),\ldots,P(A_n)$ and signatures $F(A_1),\ldots,F(A_n)$ respectively, and where $pr_i(A_i) = (P(A_i), F(A_i))$, $i = \overline{1,n}$ are projections of objects $A_1,\ldots,A_n$, which consist of properties and methods typical only for these objects.*

## Universal Operation on Objects and Runtime Class Generation

One of the most important aspects of OOP is an opportunity to work only with classes, which are described before program execution. Of course, some programs provide us with a possibility to work with new classes, which we can obtained using basic classes. However, we cannot obtain new classes, which are not inheritors of basic classes, during program execution. In modern programming, this process is known as runtime class generation (RCG) or runtime class creation (RCC). Nowadays, there are some approaches for implementation of this task for some OOP-languages, in particular for Java [24] and C# [25]. These tools based on manipulating with bytecode and implemented for such platforms of programming as Java and .NET. However, now we will not discuss practical implementations of RCG for concrete platforms of programming, but will focus on specifics of this process.

It is known, that logic of program provides access and work with some classes of objects, during program execution, depending on particular program scenario. Moreover, it can provide RCG, based on some constructors of classes. According to this, we will propose some constructors of classes, defining some universal operations on objects.

As it was mentioned before, in OOP, objects have methods. Usually they are functions, which we can execute for these objects. We can divide methods of objects on two types, depending on character of their action. They are *modifiers* and *exploiters*. Modifiers are functions, which can change objects, in particular some fields of objects. Exploiters are functions, which use objects as arguments and cannot change them. However, majority of methods of objects are local with respect to objects, and cannot be applied to objects of different types. Of course, there are some methods, which we can apply to objects of different types, but usually we need to use overloading operator for this. Nevertheless, we will define some universal operations for objects, which can be applied to any objects. Let us do it using the concept of object and object’s class.

**Definition 15.** *Union* $\cup$ *of* $n \geq 2$ *arbitrary objects is a new set of objects* $S$*, which obtain in the following way* $S = A_1 / T(A_1) \cup \ldots \cup A_n / T(A_n) = \{A_1, \ldots, A_n\} / T(S)$*, where* $A_1, \ldots, A_n$ *are objects, such that* $\forall A_i, A_j$*, where* $i, j = \overline{1,n}$ *and* $i \neq j$*,* $Eq(A_i, A_j) = 0$*,* $T(A_i)$*,* $i = \overline{1,n}$ *is a class of object* $A_i$ *and* $T(S)$ *is a class of new set of objects* $S$ *and* $n$ *is its cardinality.*

Let us consider such geometrical objects as triangle, square and trapeze. It is obvious that these objects belong to different classes of geometrical figures. Let us denote triangle as $A$, square as $B$, trapeze as $C$, and describe their classes as follows $P(A) = (p_1(A), \ldots, p_4(A))$, $P(B) = (p_1(B), \ldots, p_3(B))$, $P(C) = (p_1(C), \ldots, p_4(C))$, $F(A) = (f_1(A), f_2(A))$, $F(B) = (f_1(B), f_2(B))$ and $F(C) = (f_1(C), f_2(C))$. Properties $p_1(A)$, $p_1(B)$, $p_1(C)$ are quantities of sides of figures, properties $p_2(A)$, $p_2(B)$, $p_2(C)$, are sizes of sides of figures, properties $p_3(A)$, $p_3(B)$, are sizes of angles of figures, property $p_4(A)$ is triangle inequality and property $p_4(C)$ is parallelism of two sides of figure. Methods $f_1(A)$, $f_1(B)$, $f_1(C)$ are functions of perimeter calculation of figures, and methods $f_2(A)$, $f_2(B)$, $f_2(C)$ are functions of area calculation of figures. Of course, specifications and signatures of these objects can include more properties and methods, than we presented in this example, everything depends on level of detail.

Let us apply the union operation to these objects and create new set of objects.

$$S = A \cup B \cup C = \{A, B, C\} / T(S)$$

We have obtained new set of objects $S$ and new class of objects $T(S) = (Core(S), pr_1(A), pr_2(B), pr_3(C))$, where $Core(S) = (p_1(S), p_2(S), p_3(S), f_1(S))$, property $p_1(S)$ is quantity of sides of figures, property $p_2(S)$ means sizes of sides of figures, $p_3(S)$ are sizes of angles of figures, method $f_1(S)$ is a function of perimeter calculation of figures, $pr_1(A) = (p_4(A), f_2(A))$, $pr_2(B) = (f_2(B))$, $pr_3(C) = (p_4(C), f_2(C))$. Essentially, set of objects $S$ is the set of triangles of class $T(A)$, squares of class $T(B)$ and trapezes of class $T(C)$. Concerning class of objects $T(S)$, it describes three types of geometrical figures $T(A)$, $T(B)$ and $T(C)$.

**Definition 16.** *Intersection* $\cap$ *of two arbitrary objects* $A_1$ *and* $A_2$ *is a class of objects* $T(A) = (P(A), F(A))$*, where* $P(A) = \bigcup p_{i_1}(A_1) \mid Eq(p_{i_1}(A_1), p_{i_2}(A_2)) = 1$ *and* $F(A) = \bigcup f_{i_1}(A_1) \mid Eq(f_{i_1}(A_1), f_{i_2}(A_2)) = 1$. *Intersection of two arbitrary objects* $A_1$ *and* $A_2$ *does not exist, if and only if* $Eq(p_{i_1}(A_1), p_{i_2}(A_2)) = 0$ *for all* $i_1$ *and* $i_2$.

Let us calculate intersection of triangle $A$ and square $B$, which were described above.

$$A \cap B = T(A \cap B)$$

As the result we have obtained new class of objects $T(A \cap B)$, which does not contain any projections of objects, i.e. $T(A \cap B) = Core(A \cap B)$, where $Core(A \cap B) = (p_1(A \cap B), p_2(A \cap B), p_3(A \cap B), f_1(A \cap B))$, property $p_1(A \cap B)$ is quantity of sides of figure, property $p_2(A \cap B)$ means sizes of sides of figure, $p_3(A \cap B)$ are sizes of angles of figure, $f_1(A \cap B)$ is a function of perimeter calculation of figure. As we can see, class $T(A \cap B)$ is a class of objects, which describes some type of geometrical figures. However, we do not know exactly which type, even considering its specification and signature, because many of geometric figures have sides and angles. Nevertheless, it is a new type of objects, which we have obtained from intersection of two objects, which describe two determined types of geometric figures.

**Definition 17.** *The difference* $\setminus$ *of two arbitrary objects* $A_1$ *and* $A_2$ *is a class of objects* $T(A) = (P(A), F(A))$*, where* $P(A) = \bigcup p_{i_1}(A_1) \mid Eq(p_{i_1}(A_1), p_{i_2}(A_2)) = 0$ *and* $F(A) = \bigcup f_{i_1}(A_1) \mid Eq(f_{i_1}(A_1), f_{i_2}(A_2)) = 0$. *Difference of two arbitrary objects* $A_1$ *and* $A_2$ *does not exist, if and only if* $Eq(p_{i_1}(A_1), p_{i_2}(A_2)) = 1$ *for all* $i_1$ *and* $i_2$.

Let us calculate difference of triangle $A$ and trapeze $C$, which were described above.

$$A \setminus C = T(A \setminus C)$$

As the result, we have obtained new class of objects $T(A \setminus C)$, which does not contain core, i.e. $T(A \setminus C) = pr_1(A \setminus C)$, where $pr_1(A \setminus C) = (p_4(A), f_2(A))$. As in the case of intersection, we have obtained new class of objects, which describes some type of geometric figures, however, unlike the previous case we can say that this new type of geometric figures is a triangle, although it is described using smaller specification.

**Definition 18.** *The symmetrical difference* $\div$ *of two arbitrary objects* $A_1$ *and* $A_2$ *is a class of objects* $T(A) = (P(A), F(A))$, *where*

$$P(A) = \bigcup (p_{i_1}(A_1), p_{i_2}(A_2)) \mid Eq(p_{i_1}(A_1), p_{i_2}(A_2)) = 0,$$

$$F(A) = \bigcup (f_{i_1}(A_1), f_{i_2}(A_2)) \mid Eq(f_{i_1}(A_1), f_{i_2}(A_2)) = 0.$$

*Symmetrical difference of two arbitrary objects* $A_1$ *and* $A_2$ *does not exist, if and only if* $Eq(p_{i_1}(A_1), p_{i_2}(A_2)) = 1$ *for all* $i_1$ *and* $i_2$.

Let us calculate symmetrical difference using the same figures, i.e. triangle $A$ and trapeze $C$.

$$A \div C = T(A \div C)$$

As the result, we have obtained new class of objects $T(A \div C)$, which like the previous case, does not contain core, i.e. $T(A \div C) = (pr_1(A \div C), pr_2(A \div C))$, where $pr_1(A \div C) = (p_4(A), f_2(A))$ and $pr_2(A \div C) = (p_4(C), f_2(C))$. As in the case of difference, we have obtained new class of objects, which describes a type of geometric figures, however, as in the case of intersection, we do not know exactly which type of geometrical figures this class describes. Nevertheless, we can say that this class describes two types of geometrical figures, and one of them is triangle.

**Definition 19.** *Clone of the arbitrary object* $A$ *is an object* $A_i = A_i / P(A)$, *where* $P(A)$ *is specification of object* $A$ *and* $i$ *is a number of its clone.*

This operation is similar to concept of copy constructor in C++ [8], and is an example of constructor of objects.

As we can see, the majority of operations on objects that defined above are similar to set-theoretic operations in classical set theory [23]. However, in contrast to them, operations on objects give us opportunities to create sets and classes of objects.

## Sets of Objects and Inhomogeneous Classes

**Sets of Objects.** There are a few ways to create set of objects. Firstly, we can obtain a set of objects using union operation to not less than two arbitrary objects. Secondly, we can do it using set-theoretic operations over sets of objects. Eventually, we can obtain a set of objects combining these two approaches, namely to use union operation to not less than two arbitrary objects and not less than one arbitrary set of objects. Let us define concept of set of objects, using these ideas, concepts of object and class of objects.

**Definition 20.** *The set of objects* $S$ *is a union, which satisfies one of the following schemes:*

$$S1) O_1 \cup \ldots \cup O_n = S / T(S);$$

$$S2) S_1 \cup \ldots \cup S_m = S / T(S);$$

$$S3) O_1 \cup \ldots \cup O_n \cup S_1 \cup \ldots \cup S_m = S / T(S);$$

*where* $O_1, \ldots, O_n$ *are arbitrary objects,* $S_1, \ldots, S_m$ *are arbitrary sets of objects, and* $T(S)$ *is a class of new set of objects* $S$.

We have defined union operation for scheme $S1$, now let us do this for schemes $S2$ and $S3$. As we know, scheme $S2$ is defined in classical set theory, and this operation is known as a union of sets [23]. However, that definition does not consider concept of class of objects, that is why we need to redefine this operation.

**Definition 21.** *Union* $\cup$ *of* $m \geq 2$ *arbitrary sets of objects is a new set of objects* $S$, *which obtains in the following way* $S = S_1 / T(S_1) \cup \ldots \cup S_m / T(S_m) = \{A_1, \ldots, A_n\} / T(S)$, *where* $A_1, \ldots, A_n$, *are such objects that* $\forall A_i, A_j$, *where* $i, j = \overline{1,n}$ *and* $i \neq j$, $Eq(A_i, A_j) = 0$, $T(S_i)$, $i = \overline{1,m}$ *is a class of set of object* $S_i$ *and* $T(S)$ *is a class of new set of objects* $S$ *and* $n$ *is its cardinality.*

Let us consider two sets of objects $S_1 = \{A, B\}$, $S_2 = \{A, C\}$ which consist of triangle $A$, square $B$ and trapeze $C$, which were described above, and calculate their union.

$$S = S_1 / T(S_1) \cup S_2 / T(S_2) = \{A, B\} / T(S_1) \cup \{A, C\} / T(S_2) = \{A, B, C\} / T(S)$$

As we can see, we have obtained the same result, as in the case of union of objects $A$, $B$ and $C$, which we considered before.

Now, let us define union operations for scheme $S3$.

**Definition 22.** *Union* $\cup$ *of* $n \geq 1$ *arbitrary objects and* $m \geq 1$ *arbitrary sets of objects is a new set of objects* $S$, *which obtains in the following way*

$$S = A_1 / T(A_1) \cup \ldots \cup A_n / T_n \cup S_1 / T(S_1) \cup \ldots \cup S_m / T(S_m) = \{A_1, \ldots, A_k\} / T(S),$$

*where* $A_1,...,A_k$, *are such objects that* $\forall A_i, A_j$, *where* $i, j = \overline{1,k}$ *and* $i \neq j$, $Eq(A_i, A_j) = 0$, $T(A_p)$, $p = \overline{1,n}$ *is a class of object* $A_p$, $T(S_w)$, $w = \overline{1,m}$ *is a class of set of object* $S_w$ *and* $T(S)$ *is a class of new set of objects* $S$ *and* $k$ *is its cardinality.*

Let us consider objects $A$, $B$, $C$, and sets of objects $S_1$, $S_2$, which we used in the previous example, and calculate their union.

$$S = A/T(A) \cup B/T(B) \cup C/T(C) \cup S_1/T(S_1) \cup S_2/T(S_2) =$$
$$= A/T(A) \cup B/T(B) \cup C/T(C) \cup \{A, B\}/T(S_1) \cup \{A, C\}/T(S_2) = \{A, B, C\}/T(S)$$

**Multisets of Objects.** As we know, a multiset is a generalization of the notion of set in which members are allowed to appear more than once [29-31]. Formally multiset can be defined as a 2-tuple $(A, m)$, where $A$ is the set, and $m$ is the function that puts a natural number in accordance to each element of the set $A$ , which is called the multiplicity of the element, i.e. $m: A \rightarrow N$. However, this definition does not explain how to create the multiset of objects that is why we are going to define multiset of objects using concept of set of objects.

**Definition 23.** *The multiset of objects is a set of objects* $S = \{A_1,...,A_n\}$, *that* $\exists A_i, A_j$, $i, j = \overline{1,n}$ *and* $i \neq j$, $Eq(A_i, A_j) = 1$.

We can obtain multiset of objects in the same way as set of objects. The example will be shown, using objects and sets of objects, which were mentioned previously.

Union of objects.

$$S = A/T(A) \cup A/T(A) \cup B/T(B) \cup C/T(C) \cup C/T(C) \cup C/T(C) = \{A, A, B, C, C, C\}/T(S)$$

Union of sets of objects.

$$S = S_1/T(S_1) \cup S_2/T(S_2) = \{A, B\}/T(S_1) \cup \{A, C\}/T(S_2) = \{A, A, B, C\}/T(S)$$

Union of objects and sets of objects.

$$S = A/T(A) \cup C/T(C) \cup S_1/T(S_1) = A/T(A) \cup C/T(C) \cup \{A, B\}/T(S_1) = \{A, A, B, C\}/T(S)$$

As we can see, creation of sets and multisets of objects entails creation of new classes of objects, in particular inhomogeneous, that causes some problems related to description of such classes in OOP. At first glance it may seems, that there is no necessity to work with such classes, and we can use just concept of homogeneous class, that implemented within OOP. However, if we analyze some aspects of human mind, especially mechanisms of its thinking and analysis, we can conclude that concept of set of objects is one of the basic and very important concepts for them. Really, let us consider situation, when you need to find some book among big amount of different books, which lying on the shelves of bookcase. If you know how exactly this book looks, you can imagine this book and you can distinguish it from other books in this bookcase. Finally, during searching for this book, you perform certain exhaustive search, and at the same time, you create set of books, which you have checked. Let us imagine another situation, when you need to count money, which you have in your wallet. During counting, you create at least two sets, set of banknotes and set of coins. In addition, we can consider situation when you want to play chess, and after opening the box with chess, you need to make initial arrangement of figures on the chessboard. During figures placement, you create set of white and set of black figures from set of all figures. These are just a few simple examples from our daily activity. Very often in such situations and similar to them we operate with sets of objects subconsciously, and as the result, we pay little attention to this, but it is happening permanently. That is why we can conclude that creation of sets and classes of objects is normal activity of our mind, and if we want to simulate and reproduce this process, using computer system, we need to have certain opportunities to work with sets of objects and inhomogeneous classes of objects. It means we need appropriate tools within certain OOP-language for working with such abstractions.

Nowadays, there are a few different implementations of tools for working with sets within some OOP-languages, in particular set in STL for C++ [26], HashSet in Java [20], HashSet, SortedSet and ISet in C# [27], set and frozenset in Python [28]. These tools allow sets creation, executing set-theoretic operations, membership checking, adding and removing of elements and checking of equivalence between sets, etc. However, we have not opportunities for working with classes of sets and multisets, and this is one of disadvantages of modern OOP.

## Conclusions

In this paper we analyzed such basic concepts of OOP as object and class and their relation with set theory and AI. According to this, definitions of object and its properties, homogeneous and inhomogeneous classes of objects were proposed. We defined such operations on objects as union, intersection, difference, symmetric difference and cloning. Besides, we considered process of runtime class generation and showed that intersection, difference, symmetric difference of objects are examples of classes’ generators.

Process of sets and multisets creation was considered in details from different sides, in particular mathematical set theory, OOP and AI. After that, constructive definition of set of objects, multiset of objects and methods of their creation were proposed. Relation between sets of objects and classes of objects was also shown. The proposed approach allows not only to create (generate) sets and multisets of objects, but also to classify them. It also allows considering the problem of object classification and identification in another way.

Besides, we showed that, we could not operate with inhomogeneous classes of objects, classes of sets and multisets of objects using existing OOP tools. In addition, we cannot create new classes of object using operations on objects, in particular in runtime. After all, we can conclude that we need some OOP extension for the purpose of design and developing AIS using proposed approach.